\documentclass[runningheads]{llncs}
\usepackage[dvipdfmx]{graphicx}
\usepackage{amsmath,amssymb} 
\usepackage{color}
\usepackage{hyperref}
\usepackage[width=122mm,left=12mm,paperwidth=146mm,height=193mm,top=12mm,paperheight=217mm]{geometry}
\usepackage[vlined,boxed,commentsnumbered]{algorithm2e}
\usepackage{subfigure}
\usepackage{adjustbox,lipsum}
\usepackage{floatrow}
\usepackage{breakcites}
\newfloatcommand{capbtabbox}{table}[][\FBwidth]
\usepackage{blindtext}
\usepackage[absolute]{textpos}
\newcommand{\reals}[1]{\mathbb{R}^{#1}}
\newcommand{\enorm}[1]{\left\|{#1}\right\|}
\newcommand{\norm}[1]{\left\|{#1}\right\|}

\newcommand{\set}[1]{\left\{#1\right\}}

\DeclareMathOperator*{\dsp}{DSP}
\newcommand{\half}{\frac{1}{2}}

\DeclareMathOperator{\vecp}{vec}
\newcommand{\stiefel}{\mathcal{S}}

\newcommand{\eye}[1]{\mathbf{I}_{#1}}

\newcommand{\para}[1]{\noindent\paragraph*{\textbf{#1}}}

\renewcommand\cdots{...}

\newcommand{\suptensor}[1]{\mathfrak{S}^{d}}

\DeclareMathOperator*{\argmin}{arg\,min}
\newcommand{\fnorm}[1]{\left\|{#1}\right\|_F}

\newcommand{\comment}[1]{}

\newcommand{\done}[1]{}
\newcommand{\actodo}[1]{}
\newcommand{\bigoh}{O}

\newcommand{\dataset}{\mathcal{X}}

\DeclareMathOperator*{\argmax}{arg\,max}

\usepackage{lipsum}

\begin{document}


\begin{textblock}{13}(0.3,0.1)
\scriptsize {This is a revised version of the ECCV 2018 conference paper titled \newline \href{http://openaccess.thecvf.com/content_ECCV_2018/papers/Jue_Wang_Learning_Discriminative_Video_ECCV_2018_paper.pdf}{"Learning Discriminative Video Representations Using Adversarial Perturbations".}}
\end{textblock}

\title{Contrastive Video Representation Learning via Adversarial Perturbations} 

\titlerunning{Contrastive Video Representation Learning via Adversarial Perturbations}

\authorrunning{J. Wang and A. Cherian}
\author{Jue Wang\inst{1}\thanks{Work done while interning at MERL.}\and
Anoop Cherian\inst{2}}

\institute{${^1}$Data61/CSIRO, ANU, Canberra \quad ${^2}$MERL Cambridge, MA\\
	\email{jue.wang@anu.edu.au\quad cherian@merl.com}
}

\maketitle
\begin{abstract}
Adversarial perturbations are noise-like patterns that can subtly change the data, while failing an otherwise accurate classifier. In this paper, we propose to use such perturbations within a novel contrastive learning setup to build negative samples, which are then used to produce improved video representations. To this end, given a well-trained deep model for per-frame video recognition, we first generate adversarial noise adapted to this model. Positive and negative bags are produced using the original data features from the full video sequence and their perturbed counterparts, respectively. Unlike the classic contrastive learning methods, we develop a binary classification problem that learns a set of discriminative hyperplanes -- as a subspace -- that will separate the two bags from each other. This subspace is then used as a descriptor for the video, dubbed \emph{discriminative subspace pooling}. As the perturbed features belong to data classes that are likely to be confused with the original features, the discriminative subspace will characterize parts of the feature space that are more representative of the original data, and thus may provide robust video representations. To learn such descriptors, we formulate a subspace learning objective on the Stiefel manifold and resort to Riemannian optimization methods for solving it efficiently. We provide experiments on several video datasets and demonstrate state-of-the-art results.
\end{abstract}
\section{Introduction}
\label{intro}
\begin{figure}
	\begin{center}
        \includegraphics[width=0.8\linewidth,trim={0cm 0cm 0cm 0cm},clip]{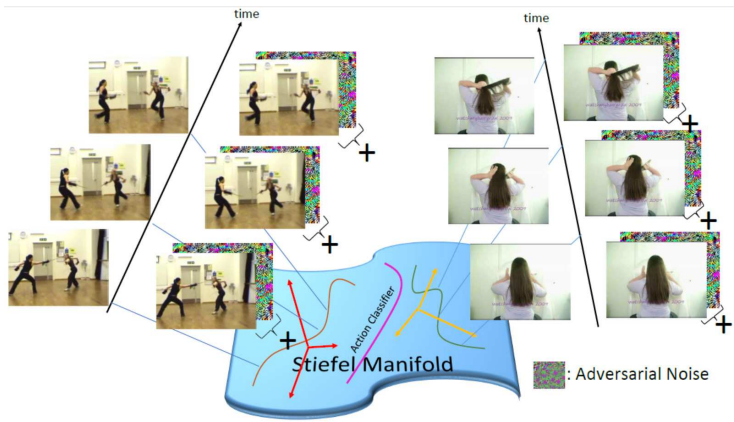}
	\end{center}
	\caption{A graphical illustration of our discriminative subspace pooling with adversarial noise. For every video sequence (as CNN features), our scheme generates a positive bag (with these features) and a negative bag by adding adversarial perturbations to the features. Next, we learn discriminative temporally-ordered hyperplanes that separate the two bags. We use orthogonality constraints on these hyperplanes and use them as representations for the video. As such representations belong to a Stiefel manifold, we use a classifier on this manifold for video recognition. }
	\label{fig:1}
\end{figure}

Deep learning has enabled significant advancements in several areas of computer vision; however, the sub-area of video-based recognition continues to be elusive. In comparison to image data, the volumetric nature of video data makes it significantly more difficult to design models that can remain within the limitations of existing hardware and the available training datasets. Typical ways to adapt  image-based deep models to videos are to resort to recurrent deep architectures or use three-dimensional spatio-temporal convolutional filters~\cite{carreira2017quo,tran2015learning,simonyan2014two}. Due to hardware limitations, the 3D filters cannot be arbitrarily long. As a result, they usually have fixed  temporal receptive fields (of a few frames)~\cite{tran2015learning}. While recurrent networks, such as LSTM and GRU, have shown promising results on video tasks~\cite{zhu2016co,liu2016spatio,ballas2015delving}, training them is often difficult, and so far their performance has been inferior to models that look at parts of the video followed by a late fusion~\cite{carreira2017quo,simonyan2013deep}.

While, better CNN architectures, such as the recent I3D framework~\cite{carreira2017quo}, is essential for pushing the state-of-the-art on video tasks, it is also important to have efficient representation learning schemes that can capture the long-term temporal video dynamics from predictions generated by a temporally local model. Recent efforts in this direction, such as rank pooling, temporal segment networks and temporal relation networks~\cite{Wang2016,grp,fernando2016learning,feichtenhofer2017temporal,fernando2015modeling,bilen2016dynamic,soo2017interpretable}, aim to incorporate temporal dynamics over clip-level features. However, such models often ignore the noise in the videos, and use representations that adhere to a plausible criteria. For example, in the  rank pooling scheme~\cite{grp,fernando2016learning,fernando2015modeling,bilen2016dynamic,fernando2016discriminative}, it is assumed that the features from each frame are temporally-ordered, and learns a representation that preserves such order -- however without accounting for whether the learned representation fits to data foreground or background. 

In this paper, we present a novel pooling framework to contrastively summarize the temporally-ordered video features. Different from prior works, we assume that per-frame video features consist of noisy parts that could confuse a classifier in a downstream task, such as for example, action recognition. A robust representation, in this setting, will be one that could avoid the classifier from using these vulnerable features for making predictions. Learning such representations is similar in motivation to the idea of contrastive learning~\cite{hadsell2006dimensionality}, which has achieved promising results in many recent works for the task of unsupervised visual feature learning~\cite{bachman2019learning,henaff2019data,hjelm2018learning,oord2018representation,tian2019contrastive,wu2018unsupervised,zhuang2019local,he2019momentum,chen2020simple}. These works learn the visual representation by contrasting the positive pairs against the negative ones via a loss function, namely as contrastive loss. However, it is challenging to generate negative samples for video sequences and building the memory bank is also not appealing. To this end, we resort to some intuitions made in a few works recently in the area of adversarial perturbations~\cite{lu2017safetynet,oh2017adversarial,moosavi2017universal,xie2017adversarial}. Such perturbations are noise-like patterns that, when added to data, can fail an otherwise well-trained highly accurate classifier. Such perturbations are usually subtle, and in image recognition tasks, are quasi-imperceptible to a human. It was shown in several recent works that such noise can be learned from data. Specifically, by taking gradient ascent on a minimizing learning objective, one can produce such perturbations that will push the data points to the class boundaries, thereby making the classifier to mis-classify. Given that the strength (norm) of this noise is often bounded, it is highly likely that such noise will find minimum strength patterns that select features that are most susceptible to mis-classification. To this end, we use the recent universal adversarial perturbation generation scheme \cite{moosavi2017universal}. 

Once the perturbations are learned (and fixed) for the dataset, we use it to learn robust representations for the video. To this end, for features from every frame, we make two bags, one consisting of the original features, while the other one consisting of features perturbed by noise. Next, we learn a discriminative hyperplane that separates the bags in a max-margin framework. Such a hyperplane, which in our case is produced by a primal support vector machine (SVM), finds decision boundaries that could well-separate the bags; the resulting hyperplane is a single vector and is a weighted combination of all the data points in the bags. Given that the data features are non-linear, and given that a kernelized SVM might not scale well with sequence lengths, we propose to instead use multiple hyperplanes for the classification task, by stacking several such hyperplanes into a column matrix. We propose to use this matrix as our data representation for the video sequence. 

However, there is a practical problem with our descriptor; each such descriptor is local to its respective sequences and thus may not be comparable between videos. To this end, we make additional restrictions on the hyperplanes -- regularizing them to be orthogonal, resulting in our representation being subspaces. Such subspaces mathematically belong to the so-called Stiefel manifold~\cite{boothby1986introduction}. We formulate a novel objective on this manifold for learning such subspaces on video features.  Further, as each feature is not independent of the previous ones, we make additional temporal constraints. We provide efficient Riemannian optimization algorithms for solving our objective, specifically using the Riemannian conjugate gradient scheme that has been used in several other recent works~\cite{grp,harandi2014manifold,huang2015projection}. Our overall pipeline is graphically illustrated in Figure~\ref{fig:1}.

We present experiments on three video recognition tasks, namely (i) action recognition, (ii) dynamic texture recognition, and (iii) 3D skeleton based action recognition. On all the experiments, we show that our scheme leads to state-of-the-art results, often improving the accuracy between 3--14\%.

Before moving on, we summarize the main contributions of this work:
\begin{itemize}
\item We introduce adversarial perturbations into the video recognition setting for contrastively learning robust video representations. 
\item We formulate a binary classification problem to learn temporally-ordered discriminative subspaces that separate the data features from their perturbed counterparts.
\item We provide efficient Riemannian optimization schemes for solving our objective on the Stiefel manifold.
\item Our experiments on three datasets demonstrate state-of-the-art results.
\end{itemize}

\section{Related work}
\label{related_work}
Traditional video learning methods use hand-crafted features (from a few frames) -- such as dense trajectories, HOG, HOF, etc.~\cite{wang2013action} -- to capture the appearance and the video dynamics, and summarize them using a bag-of-words representation or more elegantly using Fisher vectors~\cite{sadanand2012action}. With the success of deep learning methods, feeding video data as RGB frames, optical flow subsequences, RGB differences, or 3D skeleton data directly into CNNs is preferred. 
One successful such approach is the two-stream model (and its variants)~\cite{simonyan2014two,feichtenhofer2017temporal,feichtenhofer2017spatiotemporal,hayat2015deep} that use video segments (of a few frames) to train deep models, the predictions from the segments are fused via average pooling to generate a video level prediction. There are also extensions of this approach that directly learn models in an end-to-end manner~\cite{feichtenhofer2017spatiotemporal}. While, such models are appealing to capture the video dynamics, it demands memory for storing the intermediate feature maps of the entire sequence, which may be impractical for long sequences. Recurrent models~\cite{baccouche2011sequential,donahue2015long,du2015hierarchical,li2016action,srivastava2015unsupervised,yue2015beyond} have been explored for solving this issue, that can learn to filter useful information while streaming the videos through them, but they are often found difficult to train~\cite{pascanu2013difficulty}; perhaps due to the need to back-propagate over time. Using 3D convolutional kernels~\cite{carreira2017quo,tran2015learning} is another idea that proves to be promising, but bring along more parameters. The above architectures are usually trained for improving the classification accuracy, however, do not consider the robustness of their internal representations -- accounting for which may improve their generalizability to unseen test data. 

One way to improve the data representation is via contrastive learning~\cite{hadsell2006dimensionality}, which learns representations by minimizing a contrastive loss between the positive and negative pairs. This approach has been used in several recent works~\cite{bachman2019learning,henaff2019data,hjelm2018learning,oord2018representation,tian2019contrastive,wu2018unsupervised,zhuang2019local,he2019momentum,chen2020simple}, achieving promising results for unsupervised visual representation learning. Although their motivations are different, the core idea is to unsupervisely train an encoder by minimizing the contrastive loss, which encodes a visual representation closer to its positive data points and far away from its negatives. The difference with our formulation is that some works~\cite{bachman2019learning,henaff2019data,hjelm2018learning,oord2018representation,tian2019contrastive,zhuang2019local,chen2020simple} formulate the positive and negative pairs within a mini-batch while others~\cite{he2019momentum,wu2018unsupervised} build a memory bank for generating the pairs. However, as the video data is often voluminous, it is hard to apply the same strategy for learning video representations. Moreover, the size of the memory bank could be huge due to the potentially large spatio-temporal semantic complexity in the video data. Instead, we show how we can use  adversarial perturbations~\cite{moosavi2017universal} to produce negative samples, in a network-agnostic manner, which can then be used for contrastive learning within a novel subspace-based contrastive learning framework. Specifically, different from the classic contrastive learning methods mentioned above, we formulate a binary classification problem that contrasts the video features against its perturbed counterparts and use the learned decision boundaries as video representation.

Our main inspiration comes from the recent work of Moosavi et al.~\cite{moosavi2017universal} that show the existence of quasi-imperceptible image perturbations that can fool a well-trained CNN model. They provide a systematic procedure to learn such perturbations in an image-agnostic way. In Xie et al.~\cite{xie2017adversarial}, such perturbations are used to improve the robustness of an object detection system. Similar ideas have been explored in ~\cite{lu2017safetynet,oh2017adversarial,zhang2018deep}. In Sun et al.~\cite{sun2014discover}, a latent model is used to explicitly localize discriminative video segments. In Chang et al.~\cite{chang2017semantic}, a semantic pooling scheme is introduced for localizing events in untrimmed videos. While these schemes share similar motivation as ours, the problem setup and formulations are entirely different. 

On the representation learning front of our contribution, there are a few prior pooling schemes that are similar in the sense that they also use the parameters of an optimization functional as a representation. The most related work is rank-pooling and its variants~\cite{fernando2016learning,fernando2015modeling,fernando2016discriminative,su2016hierarchical,bilen2017action,cherian2018non,wang2017ordered} that use a rank-SVM for capturing the video temporal evolution. Similar to ours, Cherian et al.~\cite{grp} propose to use a subspace  to represent video sequences. However, none of these methods ensure if the temporal-ordering constraints capture useful video content or capture some temporally-varying noise. To overcome this issue, Wang et al~\cite{wang2018video} proposes a contrastive video representation learning scheme using the decision boundaries of a support vector machine classifier that contrast data features against independently sampled noise. In this paper, we revisit this problem in the setting of data dependent noise generation via an adversarial noise design and learns a non-linear decision boundary using Riemannian optimization; our learned representations per sequence are more expressive and leads to significant performance benefits. 

\section{Proposed Method}
Let us assume $X=\left<x_1, x_2, \cdots, x_n\right>$ be a sequence of video features, where $x_i\in\reals{d}$ represents the feature from the $i$-th frame. We use `frame' in a loose sense; it could mean a single RGB frame or a sequence of a few RGB or optical flow frames (as in the two stream~\cite{simonyan2014very} or the I3D architectures~\cite{carreira2017quo}) or a 3D skeleton. The feature representation $x_i$ could be the outputs from intermediate layers of a CNN. As alluded to in the introduction, our key idea is the following. We look forward to an effective representation of $X$ that is (i) compact, (ii) preserves characteristics that are beneficial for the downstream task (such as video dynamics), and (iii) efficient to compute. Recent methods such as generalized rank pooling~\cite{grp} have similar motivations and propose a formulation that learns compact temporal descriptors that are closer to the original data in $\ell_2$ norm. However, such a reconstructive objective may also capture noise, thus leading to sub-optimal performance. Instead, we take a different approach in the contrastive learning fashion. Specifically, we assume to have access to some noise features $Z=\set{z_1, z_2,\cdots, z_m}$, each $z_i\in\reals{d}$. Let us call $X$ the positive bag, with a label $y=+1$ and $Z$ the negative bag with label $y=-1$. Our main goal is to find a discriminative hyperplane that separates the two bags; these hyperplanes can then be used as the representation for the bags. 

An obvious question is how such a hyperplane can be a good data representation? To answer this, let us consider the following standard SVM formulation with a single discriminator $w\in\reals{d}$:
\begin{equation}
\min_{w, \xi\geq 0} \half\enorm{w}^2 + \sum_{\theta\in X\cup Z} \left[\max(0, 1- y(\theta) w^\top \theta + \xi_{\theta}) + C \xi_{\theta}\right],
\label{eq:1}
\end{equation}
where with a slight abuse of notation, we assume $y(\theta) \in \set{+1, -1}$ is the label of $\theta$, $\xi$ are the slack variables, and $C$ is a regularization constant on the slacks. Given the positive and negative bags, the above objective learns a linear classification boundary that could separate the two bags with a classification accuracy of say $\gamma$. If the two bags are easily separable, then the number of support vectors used to construct the separating hyperplane might be a few and thus may not capture a weighted combination of a majority of the points in the bags –- as a result, the learned hyperplane would not be representative of the bags. However, if the negative bag $Z$ is suitably selected and we demand a high $\gamma$, we may turn~\eqref{eq:1} into a difficult optimization problem and would demand the solver to overfit the decision boundary to the bags; this overfitting creates a significantly better summarized representation, as it may need to span a larger portion of the bags to satisfy the $\gamma$ accuracy.\footnote{Here regularization parameter $C$ is mainly assumed to help avoid outliers.} This overfitting of the hyperplane is our key idea, that allows to avoid using data features that are susceptible to perturbations, while summarizing the rest.

There are two key challenges to be addressed in developing such a representation, namely (i) an appropriate noise distribution for the negative bag, and (ii) a formulation to learn the separating hyperplanes. We explore and address these challenges below.
\begin{algorithm} [t]
	\SetAlgoLined
	\KwIn{Feature points $x_{ij}$, Network weighting $W$, fooling rate $\psi$, cross entropy loss with softmax funtion $f(.)$, normalization operator $N(.)$.}
    \KwOut{Adversarial noise vector $\epsilon$.}
	Initialization: $\epsilon\gets 0.$ \\
	\Repeat{$Accuracy\leq 1-\psi$} {
		$\Delta \epsilon\gets \argmin_{r} \|r\|_2 - \sum_{ij} f(W^\top (x_{ij}), W^\top (x_{ij}+\epsilon+r))$\;
		$\epsilon \gets N(\epsilon + \Delta \epsilon$)\;
	}
	\KwRet{$v$}
	\caption{Optimization step for solving adversarial noise.}
	\label{alg:1}
\end{algorithm}

\subsection{Finding Noise Patterns}
As alluded to above, having good noise distributions that help us identify the vulnerable parts of the feature space is important for our scheme to perform well. The classic contrastive learning schemes either pair random noise~\cite{tian2019contrastive,wang2018video} or use in-batch samples~\cite{chen2020simple,he2019momentum} as the negatives, in which the random noise may introduce uncertainty into the learned feature and the in-batch samples would take too much memory during the training of video data. Instead, we resort to the recent idea of universal adversarial perturbations (UAP)~\cite{moosavi2017universal} to formulate the negatives by adding this global noise pattern onto the positives. This scheme is dataset-agnostic and provides a systematic and mathematically grounded formulation for generating adversarial noise that when added to the original features is highly-likely to mis-classify a pre-trained classifier. Further, this scheme is computationally efficient and requires less data for building relatively generalizable universal perturbations.

Precisely, suppose $\dataset$ denotes our dataset, let $h$ be a CNN trained on $\dataset$ such that $h(x)$ for $x\in\dataset$ is a class label predicted by $h$. Universal perturbations are noise vectors $\epsilon$ found by solving the following objective:
\begin{equation}
\min_{\epsilon} \norm{\epsilon} \text{ s.t. } h(x+\epsilon) \neq h(x), \forall x \in \dataset,
\label{eq:uap}
\end{equation}
where $\norm{\epsilon}$ is a suitable normalization on $\epsilon$ such that its magnitude remains small, and thus will not change $x$ significantly. In~\cite{moosavi2017universal}, it is argued that this norm-bound restricts the optimization problem in~\eqref{eq:uap} to look for the minimal perturbation $\epsilon$ that will move the data points towards the class boundaries; i.e., selecting features that are most vulnerable -- which is precisely the type of noise we need in our representation learning framework. 


To this end, we extend the scheme described in~\cite{moosavi2017universal}, to our \textcolor{red}{contrastive setting}. Differently to their work, we aim to learn a UAP on high-level CNN features as detailed in Alg.~\ref{alg:1} above, where the $x_{ij}$ refers to the $i^{th}$ frame in the $j^{th}$ video. We use the classification accuracy before and after adding the noise as our optimization criteria as captured by maximizing the cross-entropy loss.

\subsection{Discriminative Subspace Pooling}
Once a ``challenging'' noise distribution is chosen, the next step is to find a summarization technique for the given video features. While one could use a simple discriminative classifier, such as described in~\eqref{eq:1} to achieve this, such a linear classifier might not be sufficiently powerful to separate the potentially non-linear CNN features and their perturbed counterpart. An alternative is to resort to non-linear decision boundaries using a kernelized SVM; however that may make our approach less scalable and poses challenges for end-to-end learning. Thus, we look forward to a representation within the span of data features, while having more capacity for separating non-linear features.  

Our main idea is to use a subspace of discriminative directions (as against a single one as in~\eqref{eq:1}) for separating the two bags such that every feature $x_i$ is classified by at least one of the hyperplanes to the correct class label. Such a scheme can be looked upon as an approximation to a non-linear decision boundary by a set of linear ones, each one separating portions of the data. Mathematically, suppose $W\in\reals{d\times p}$ is a matrix with each hyperplane as its columns, then we seek to optimize:
\begin{equation}
\min_{W, \xi} \Omega(W) + \sum_{\theta \in X\cup Z} \left[\max\left(0, 1-\max\left(\bf{y}(\theta) \odot W^\top \theta\right) - \xi_{\theta}\right) + C\xi_{\theta}\right],
\label{eq:2}
\end{equation}
where $\bf{y}$ is a vector with the label $y$ repeated $p$ times along its rows. The quantity $\Omega$ is a suitable regularization for $W$, of which one possibility is to use $\Omega(W) = W^\top W = \eye{p}$, in which case $W$ spans a $p$ dimensional subspace of $\reals{d}$. Enforcing such subspace constraints (orthonormality) on these hyperplanes are often empirically seen to demonstrate better performance as is also observed in~\cite{grp}. The operator $\odot$ is the element-wise multiplication and the quantity $\max(\bf{y}(\theta) \odot W^\top \theta)$ captures the maximum value of the element-wise multiplication, signifying that if at least one hyperplane classifies $\theta$ correctly, then the hinge-loss will be zero. 

Recall that we work with video data, and thus there are temporal characteristics of this data modality that may need to be captured by our representation. In fact, recent works show that such temporal ordering constraints indeed results in better performance, e.g., in action recognition~\cite{grp,fernando2015modeling,bilen2016dynamic,bilen2017action}.  However, one well-known issue with such ordered pooling techniques is that they impose a global temporal order on all frames jointly. Such holistic ordering ignores the repetitive nature of human actions, for example, in actions such as clapping or hand-waving. As a result, it may lead the pooled descriptor to overfit to non-repetitive features in the video data, which might be corresponding to noise/background. Usually a slack variable is introduced in the optimization to handle such repetitions, however its effectiveness is questionable. To this end, we propose a simple temporal segmentation based ordering constraints, where we first segment a video sequence into multiple non-overlapping temporal segments $\mathcal{T}_0, \mathcal{T}_1,\cdots \mathcal{T}_{\lfloor n/\delta\rfloor}$, and then enforce ordering constraints only within the segments. We find the segment length $\delta$ as the minimum number of consecutive frames that do not result in a repeat in the action features.

With the subspace constraints on $W$ and introducing temporal segment-based ordering constraints on the video features, our complete \textbf{order-constrained discriminative subspace pooling optimization} can be written as:
\begin{align}
\label{eq:5}\min_{\substack{W^\top W = \eye{p},\\\xi,\zeta\geq 0}}&\sum_{\theta \in X\cup Z}\!\!\!\!\!\left[\max\!\left(0, 1-\max\left(\bf{y}(\theta) \odot W^\top \theta\right) - \xi_{\theta}\right)\right]\!\!+\!C_1\!\!\!\!\!\sum_{\theta\in X \cup Z}\!\!\!\!\!\xi_{\theta} +\!C_2\!\!\sum_{i<j} \zeta_{ij},\\
\label{eq:6}&\enorm{W^\top x_i}^2 + 1 \leq \enorm{W^\top x_j}^2 +\zeta_{ij}, \quad i<j, \forall (i,j)\in \mathcal{T}_k, \text{where}\\
\mathcal{T}_k &= \set{k\delta+1, k\delta+2,\cdots, \min(n,(k+1)\delta)}, \forall k\in\set{0,1,\cdots, \lfloor n/\delta\rfloor} \\
\delta & = b^*-a^*, \text{ where } (a^*,b^*) = \argmin_{a,b>a} \enorm{x_a - x_b},
\label{eq:3}
\end{align}
where~\eqref{eq:6} captures the temporal order, while the last two equations define the temporal segments, and computes the appropriate segment length $\delta$, respectively. Note that, the temporal segmentation part could be done offline, by using all videos in the dataset, and selecting a $\delta$ which is the mean. In the next section, we present a scheme for optimizing $W$ by solving the objective in~\eqref{eq:5}and~\eqref{eq:6}. 

Once each video sequence is encoded by a subspace descriptor, we use a classifier on the Stiefel manifold for recognition. Specifically, we use the standard exponential projection metric kernel~\cite{grp,harandi2014expanding} to capture the similarity between two such representations, which are then classified using a kernelized SVM.

\subsection{Efficient Optimization}
The orthogonality constraints on $W$ results in a non-convex optimization problem that may seem difficult to solve at first glance. However, note that such subspaces belong to well-studied objects in differential geometry. Specifically, they are elements of the Stiefel manifold $\stiefel(d,p)$ ($p$ subspaces in $\reals{d}$), which are a type of Riemannian manifolds with positive curvature~\cite{boothby1986introduction}. There exists several well-known optimization techniques for solving objectives defined on this manifold~\cite{absil2009optimization}, one efficient scheme is Riemannian conjugate gradient (RCG)~\cite{smith1994optimization}. This method is similar to the conjugate gradient scheme in Euclidean spaces, except that in the case of curved-manifold-valued objects, the gradients should adhere to the geometry (curvature) of the manifold (such as orthogonal columns in our case), which can be achieved via suitable projection operations (called exponential maps). However, such projections may be costly. Fortunately, there are well-known approximate projection methods, termed~\emph{retractions} that could achieve these projections efficiently without losing on the accuracy. Thus, tying up all together, for using RCG on our problem, the only part that we need to derive is the Euclidean gradient of our objective with respect to $W$. To this end, rewriting~\eqref{eq:6} as a hinge loss on~\eqref{eq:5}, our objective on $W$ and its gradient are:
\begin{align}
&\min_{W\in\stiefel(d,p)} g(W):= \sum_{\theta \in X\cup Z}\left[\max\left(0, 1-\max\left(\bf{y}(\theta) \odot W^\top \theta\right) - \xi_{\theta}\right)\right] \notag\\ 
&\qquad\qquad\qquad + \frac{1}{n(n-1)}\sum_{i<j}\max(0, 1+\enorm{W^\top x_i}^2-\enorm{W^\top x_j}^2-\zeta_{ij}),
\end{align}
\begin{align}
&\frac{\partial{g}}{\partial W} = \sum_{\theta \in X\cup Z} A(W; \theta,y(\theta)) + \frac{1}{n(n-1)}\sum_{i<j}B(W;x_i,x_j), \text{where }\\
&A(W; \theta, y(\theta)) = \left\{ 
\begin{array}{ll} 
&0, \quad \text{if } \max(y(\theta)\odot W^\top\theta-\xi_{\theta}) \geq 1\\
&-\left[\mathbf{0}_{d\times r\!-\!1}\ y(\theta)\theta\  \mathbf{0}_{d\times p-r}\right],\ r=\argmax_q y(\theta)\!\odot\!W^\top_q\theta, \text{ else}\\
\end{array}
\right.\\
&B(W; x_i, x_j) = \left\{ 
\begin{array}{ll} 
&0, \quad \text{if } \enorm{W^\top x_j}^2 \geq 1+\enorm{W^\top x_i}^2 -\zeta_{ij} \\
&2(x_ix_i^\top -x_jx_j^\top)W,\quad \text{else.}\\
\end{array}
\right.
\end{align}
In the definition of $A(W)$, we use $W^\top_q$ to denote the $q$-th column of $W$. To reduce clutter in the derivations, we have avoided including the terms using $\mathcal{T}$. Assuming the matrices of the form $xx^T$ can be computed offline, on careful scrutiny we see that the cost of gradient computations on each data pair is only $\bigoh(d^2p)$ for $B(W)$ and $\bigoh(dp)$ for the discriminative part $A(W)$. If we include temporal segmentation with $k$ segments, the complexity for $B(W)$ is $\bigoh(d^2p/k)$.

\noindent\paragraph*{\textbf{End-to-End Learning:}} The proposed scheme can be used in an end-to-end CNN learning setup where the representations can be learned jointly with the CNN weights. In this case, CNN backpropogation would need gradients with respect to the solutions of an argmin problem defined in~\eqref{eq:5}, which may seem difficult. However, there exist well-founded techniques~\cite{chiang1984fundamental},~\cite{faugeras1993three}[Chapter 5] to address such problems, specifically in the CNN setting~\cite{gould2016differentiating} and such techniques can be directly applied to our setup. However, since gradient derivations using these techniques will require review of some well-known theoretical results that could be a digression from the course of this paper, we provide them in the supplementary materials. 

\section{Experiments}
\label{experiment}
In this section, we demonstrate the utility of our discriminative subspace pooling (DSP) on several standard vision tasks (including action recognition, skeleton-based video classification, and dynamic video understanding), and on diverse CNN architectures such as ResNet-152, Temporal Convolutional Network (TCN), and Inception-ResNet-v2.  We implement our pooling scheme using the ManOpt Matlab package~\cite{boumal2014manopt} and use the RCG optimizer with the Hestenes-Stiefel's~\cite{hager2005new} update rule. We found that the optimization produces useful representations in about 50 iterations and takes about 5 milli-seconds per frame on a single core 2.6GHz CPU.  We set the slack regularization constant $C=1$. As for the CNN features, we used public code for the respective architectures to extract the features. Generating the adversarial perturbation plays a key role in our algorithm, as it is used to generate our negative bag for learning the discriminative hyperplanes. We follow the experimental setting in~\cite{moosavi2017universal} to generate UAP noise for each model by solving the energy function as depicted in Alg.~\ref{alg:1}. Differently from ~\cite{moosavi2017universal}, we generate the perturbation in the shape of the high level CNN feature instead of an RGB image. We review below our the datasets, their evaluation protocols, the CNN features next.

\subsection{Datasets, CNN Architectures, and Feature Extraction}
\noindent\textbf{HMDB-51~\cite{kuehne2011hmdb}:} is a popular video benchmark for human action recognition, consisting of 6766 Internet videos over 51 classes; each video is about 20 -- 1000 frames. The standard evaluation protocol reports average classification accuracy on three-folds. To extract features, we train a two-stream ResNet-152 model (as in~\cite{simonyan2014two}) taking as input RGB frames (in the spatial stream) and a stack of optical flow frames (in the temporal stream). We use features from the pool5 layer of each stream as input to DSP, which are sequences of 2048D vectors.

\noindent\textbf{NTU-RGBD~\cite{shahroudy2016ntu}:} is by far the largest 3D skeleton-based video action recognition dataset. It has 56,880 video sequences across 60 classes, 40 subjects, and 80 views. The videos have on average 70 frames and consist of people performing various actions; each frame annotated for 25 3D human skeletal keypoints (some videos have multiple people). According to different subjects and  camera views, two evaluation protocols are used, namely cross-view and cross-subject evaluation~\cite{shahroudy2016ntu}. We use the scheme in Shahroudy et al.~\cite{shahroudy2016ntu} as our baseline in which a temporal CNN (with residual units) is applied on the raw skeleton data. We use the 256D features from the bottleneck layer (before their global average pooling layer) as input to our scheme. 

\noindent\textbf{YUP++ dataset~\cite{feichtenhofer2017temporal}:} is a recent dataset for dynamic video-texture understanding. It has 20 scene classes with 60 videos in each class. Importantly, half of the sequences in each class are collected by a static camera and the rest are recorded by a moving camera. The latter is divided into two sub-datasets, YUP++ stationary and YUP++ moving. As described in the~\cite{feichtenhofer2017temporal}, we apply the same 1/9 train-test ratio for evaluation. There are about 100-150 frames per sequence. We train an Inception-ResNet-v2 on the respective training set to generate the features and fine-tune a network that was pre-trained on the ImageNet dataset. In detail, we apply the 1/9 train-test ratio and follow the standard supervised training procedure of image-based tasks; following which we extract frame-level features (1536D) from the second-last fully-connected layer. 

\subsection{Parameter Analysis}
\begin{figure}[t]
	\begin{center}
		\subfigure[]{\label{subfig:1}\includegraphics[width=0.35\linewidth]{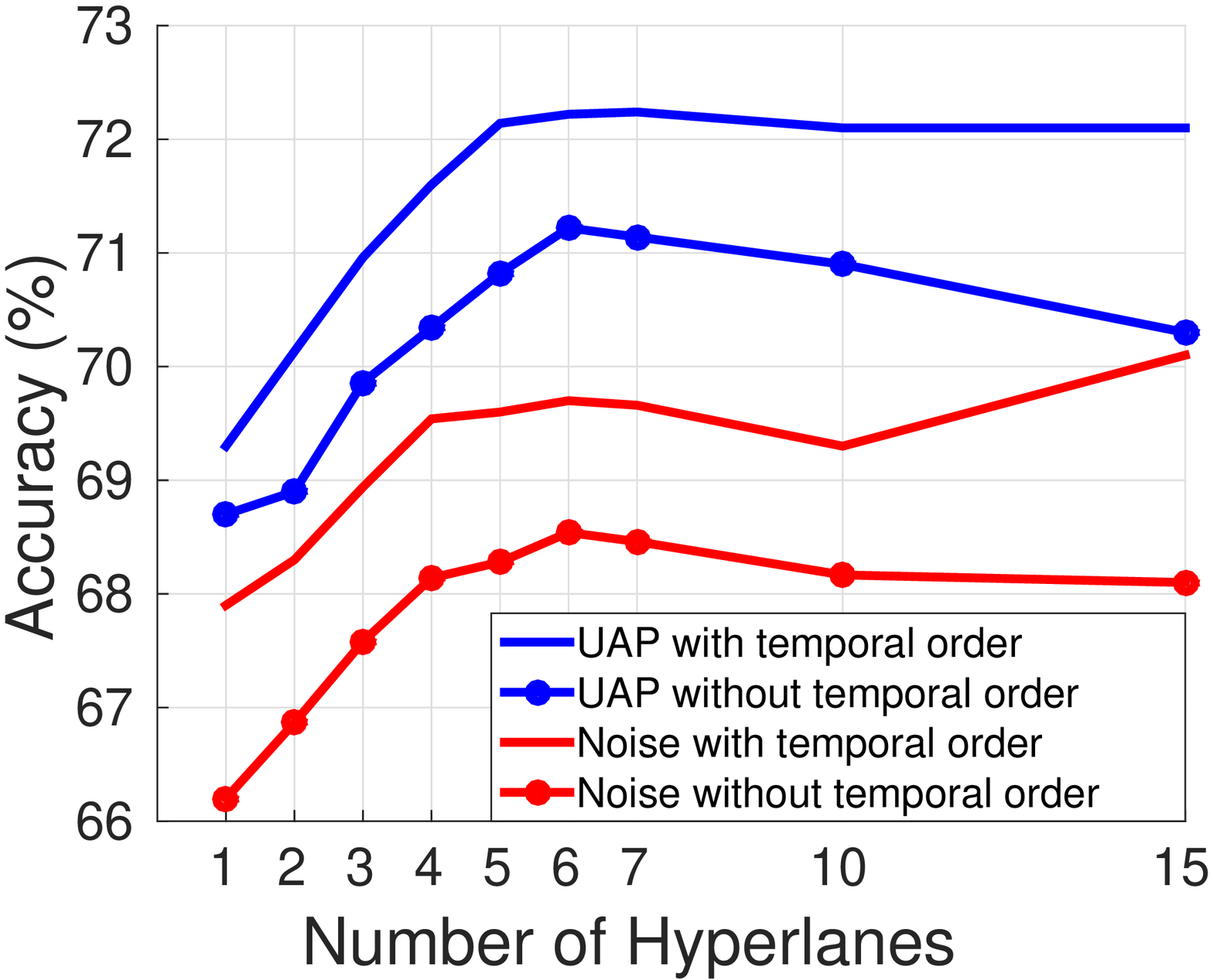}} 
        \qquad
        \subfigure[]{\label{subfig:2}\includegraphics[width=0.35\linewidth]{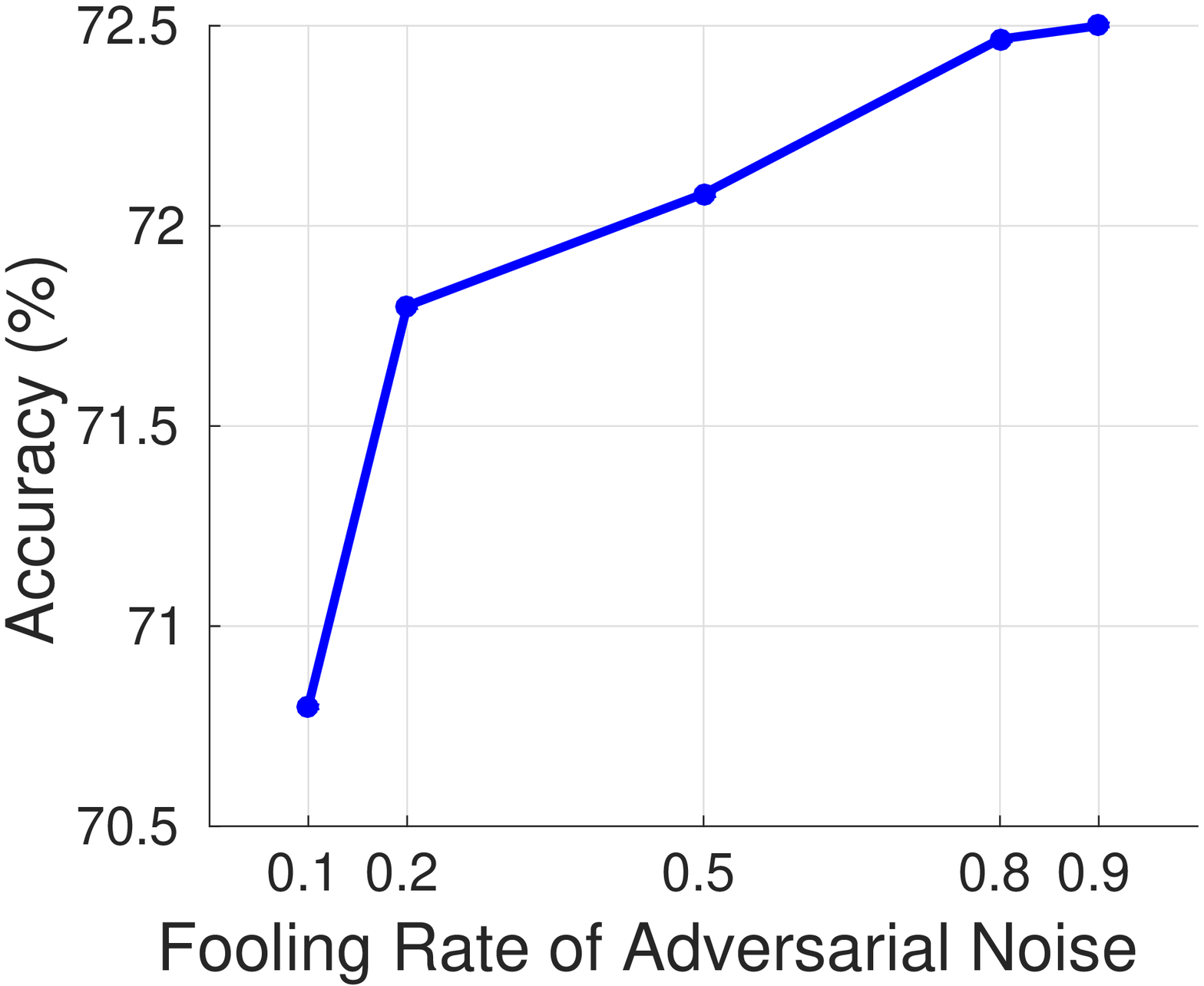}}                
	\end{center}
	\caption{Analysis of the hyper parameters used in our scheme. All experiments use ResNet-152 features on HMDB-51 split-1 with a fooling rate of 0.8 in (a) and 6 hyperplanes in (b). See text for details.}
\end{figure}
\para{Evaluating the Choice of Noise:} As is clear by now, the noise patterns should be properly chosen in the contrastive learning setup, as it will affect how well the discriminative hyperplanes characterize useful video features. To investigate the quality of UAP features, we compare it with the baseline of choosing noise from a Gaussian distribution with the data mean and standard deviation computed on the respective video dataset (as done in the work of Wang et al.~\cite{wang2018video}). We repeat this experiment 10-times on the HMDB-51 split-1 features. In Figure~\ref{subfig:1}, we plot the average classification accuracy after our pooling operation against an increasing number of hyperplanes in the subspaces. As is clear, using UAP significantly improves the performance against the alternative, substantiating our intuition. Further, we also find that using more hyperplanes is beneficial, suggesting that adding UAP to the features leads to a non-linear problem requiring more than a single discriminator to capture the informative content.

\para{Evaluating Temporal Constraints:} Next, we evaluate the merit of including temporal-ordering constraints in the DSP objective, viz.~\eqref{eq:5}. In Figure~\ref{subfig:1}, we plot the accuracy with and without such temporal order, using the same settings as in the above experiment. As is clear, embedding temporal constraint will help the discriminative subspace capture representations that are related to the video dynamics, thereby showing better accuracy. In terms of the number of hyperplanes, the accuracy increases about $3\%$ from one hyperplane to when using six hyperplanes, and drops around $0.5\%$ from 6 hyperplanes to 15 hyperplanes, suggesting that the number of hyperplanes (6 in this case) is sufficient for representing most sequences. 

\para{UAP Fooling Rate:} In Figure~\ref{subfig:2}, we analyze the fooling rate of UAP that controls the quality of the adversary to confuse the trained classifier. The higher the fooling rate is, the more it will mix the information of the feature in different classes. As would be expected, we see that increasing the fooling rate from 0.1 to 0.9 increases the performance of our pooling scheme as well. Interestingly, our algorithm could perform relatively well without requiring a very high value of the fooling rate. From ~\cite{moosavi2017universal}, a lower fooling rate would reduce the amount of data needed for generating the adversarial noise, making their algorithm computationally cheaper. Further, comparing Figures~\ref{subfig:1} and~\ref{subfig:2}, we see that incorporating a UAP noise that has a fooling rate of even 10\% does show substantial improvements in DSP performance against using Gaussian random noise (70.8\% in Figure~\ref{subfig:2} against 69.8\% in Figure~\ref{subfig:1}).


\paragraph*{\textbf{Experimental Settings:}} Going by our observations in the above analysis, for all the experiments in the sequel, we use six subspaces in our pooling scheme, use temporal ordering constraints in our objective, and use a fooling rate of 0.8 in UAP. Further, as mentioned earlier, we use an exponential projection metric kernel~\cite{cherian2018non} for the final classification of the subspace descriptors using a kernel SVM. Results using end-to-end learning are provided in the supplementary materials.

\begin{table}[t]
\centering
\begin{tabular}{|l|l|l|l|l|l|l|l|}
\hline
    & \multicolumn{3}{l|}{HMDB-51}    & \multicolumn{2}{l|}{NTU-RGBD} & \multicolumn{2}{l|}{YUP++} \\ \hline
    & Spatial & Temporal & Two-stream & Cross-subject   & Cross-view  & Stationary     & Moving    \\ \hline
AP  & 46.7\%~\cite{feichtenhofer2016convolutional}  & 60.0\%~\cite{feichtenhofer2016convolutional}   & 63.8\%~\cite{feichtenhofer2016convolutional}     & 74.3\%~\cite{soo2017interpretable}          & 83.1\%~\cite{soo2017interpretable}      & 85.1\%         & 76.5\%    \\ \hline
MP  & 45.1\%  & 58.5\%   & 60.6\%     & 65.4\%          & 78.5\%      & 81.8\%         & 72.4\%    \\ \hline
DSP & \textbf{58.5\%}  & \textbf{67.0\%}   & \textbf{72.5\%}     & \textbf{81.6\%}          & \textbf{88.7\%}      & \textbf{95.1}\%         & \textbf{88.3}\%    \\ \hline
\end{tabular}
\caption{The accuracy comparison between our Discriminative subspace pooling (DSP) with standard Average pooling (AP) and Max pooling (MP).}
\label{table:1}
\end{table}
\comment{
\subsection{Parameter analysis}
In the Figure~\ref{subfig:2}, we present how the accuracy changes with and without embedding temporal order in different number of hyperplanes. As is clear, embedding temporal constraint will help the discriminative subspace to get a better representation. In terms of the number of hyperplanes, the accuracy increases about $3\%$ from 1 hyperplane to 6 hyperplanes, and drops around $0.5\%$ from 6 hyperplanes to 15 hyperplanes, which means a number of hyperplanes (6 in this case) is enough for capturing the discriminative feature of the sequence, redundant ones would catch the repeated information and cause confusion for the classifier. Additionally, in Figure~\ref{subfig:3}, we compare the accuracy with adversarial noise in different fooling rate. The higher the fooling rate is, the more adversarial noise will capture the information of the feature in different classes, which result in a better performance. Interestingly, our algorithm could performance relatively well without requiring a very high value of fooling rate of the adversarial noise From ~\cite{moosavi2017universal}, a lower fooling rate would reduce the amount of data that is used for generating the adversarial noise, which make the entire algorithm more computationally cheap.
}

\subsection{Experimental Results}
\paragraph*{\textbf{Compared with standard pooling:}} In Table~\ref{table:1}, we show the performance of DSP on the three datasets and compare to standard pooling methods such as average pooling and max pooling. As is clear, we outperform the baseline results by a large margin. Specifically, we achieve 9$\%$ improvement on the HMDB-51 dataset split-1 and $5\%-8\%$ improvement on the NTU-RGBD dataset. On these two datasets, we simply apply our pooling method on the CNN features extracted from the pre-trained model. We achieve a substantial boost (of up to $12\%$) after applying our scheme.

\begin{table}[t]
\centering
\scalebox{0.9}{
\begin{tabular}{lc}\hline
\multicolumn{2}{c}{HMDB-51}            \\\hline
Method                      & Accuracy \\\hline
Temporal Seg. n/w  ~\cite{Wang2016}                        & 69.4\%   \\
TS I3D ~\cite{carreira2017quo}             & 80.9\%   \\
ST-ResNet~\cite{feichtenhofer2016spatiotemporal}                    & 66.4\%   \\
ST-ResNet+IDT ~\cite{feichtenhofer2016spatiotemporal}               & 70.3\%   \\
STM Network ~\cite{feichtenhofer2017spatiotemporal}      & 68.9\%   \\
STM Network+IDT ~\cite{feichtenhofer2017spatiotemporal}      & 72.2\%   \\
ShuttleNet+MIFS ~\cite{Shi_2017_ICCV}            & 71.7\%   \\
GRP   ~\cite{grp}                      & 70.9\%   \\
SVMP ~\cite{wang2018video}                       & 71.0\%   \\
$L^2$STM~\cite{Sun_2017_ICCV}                     & 66.2\%   \\\hline
Ours(TS ResNet)     & \textbf{72.4\%}   \\
Ours(TS ResNet+IDT) & \textbf{74.3\%}   \\
Ours(TS I3D)        & \textbf{81.5\%}  \\\hline
\end{tabular}}
\quad
\scalebox{0.9}{
\begin{tabular}{lcc}\hline
\multicolumn{3}{c}{NTU-RGBD}                                                            \\\hline
Method             & \multicolumn{1}{c}{Cross-Subject} & \multicolumn{1}{c}{Cross-View} \\\hline
VA-LSTM~\cite{Zhang_2017_ICCV}             & 79.4\%                            & 87.6\%                         \\
TS-LSTM ~\cite{Lee_2017_ICCV}             & 74.6\%                            & 81.3\%                         \\
ST-LSTM+Trust Gate~\cite{liu2017skeleton} & 69.2\%                            & 77.7\%                         \\
SVMP~\cite{wang2018video}               & 78.5\%                            & 86.4\%                         \\
GRP   ~\cite{grp}             & 76.0\%                            & 85.1\%                         \\
Res-TCN~\cite{soo2017interpretable}            & 74.3\%                            & 83.1\%                         \\\hline
Ours               & \textbf{81.6\%}                   & \textbf{88.7\%}                \\\hline
\multicolumn{3}{c}{YUP++}                                                               \\\hline
Method             & \multicolumn{1}{c}{Stationary}    & \multicolumn{1}{c}{Moving}     \\\hline
TRN~\cite{feichtenhofer2017temporal}                  & 92.4\%                            & 81.5\%                         \\
SVMP~\cite{wang2018video}               & 92.5\%                            & 83.1\%                         \\
GRP~\cite{grp}                 & 92.9\%                            & 83.6\%                         \\\hline
Ours               & \textbf{95.1\%}                   & \textbf{88.3\%}                \\\hline
\end{tabular}}
\caption{Comparisons to the state-of-the-art on each dataset following their respective official evaluation protocols. We used three splits for HMDB-51. `TS' refers to `Two-Stream'.}
\label{table:2}
\end{table}

\paragraph*{\textbf{Comparisons to the State of the Art:}} In Table~\ref{table:2}, we compare DSP to the state-of-the-art results on each dataset. On the HMDB-51 dataset, we also report accuracy when DSP is combined hand-crafted features (computed using dense trajectories~\cite{wang2013dense} and summarized as Fisher vectors (IDT-FV)). As the results show, our scheme achieves significant improvements over the state of the art. For example, without IDT-FV, our scheme is 3\% better than than the next best scheme~\cite{Wang2016} (69.4\% vs. 72.4\% ours). Incorporating IDT-FV improves this to 74.3\% which is again better than other schemes. We note that the I3D architecture~\cite{carreira2017quo} was introduced recently that is pre-trained on the larger Kinectics dataset and when fine-tuned on the HMDB-51 leads to about 80.9\% accuracy. To understand the advantages of DSP on pooling I3D model generated features, we applied our scheme to their bottleneck features (extracted using the public code provided by the authors) from the fine-tuned model. We find that our scheme further improves I3D by about 0.6\% showing that there is still room for improvement for this model. On the other two datasets, NTU-RGBD and YUP++, we find that our scheme leads to about 5--7\% and 3--6\% improvements respectively, and outperforms prior schemes based on recurrent networks and temporal relation models, suggesting that our pooling scheme  captures spatio-temporal cues much better than recurrent models. 

\begin{figure}
\begin{floatrow}
\ffigbox{%
  \includegraphics[width=0.8\linewidth,trim={0cm 0cm 0cm 0cm}, clip]{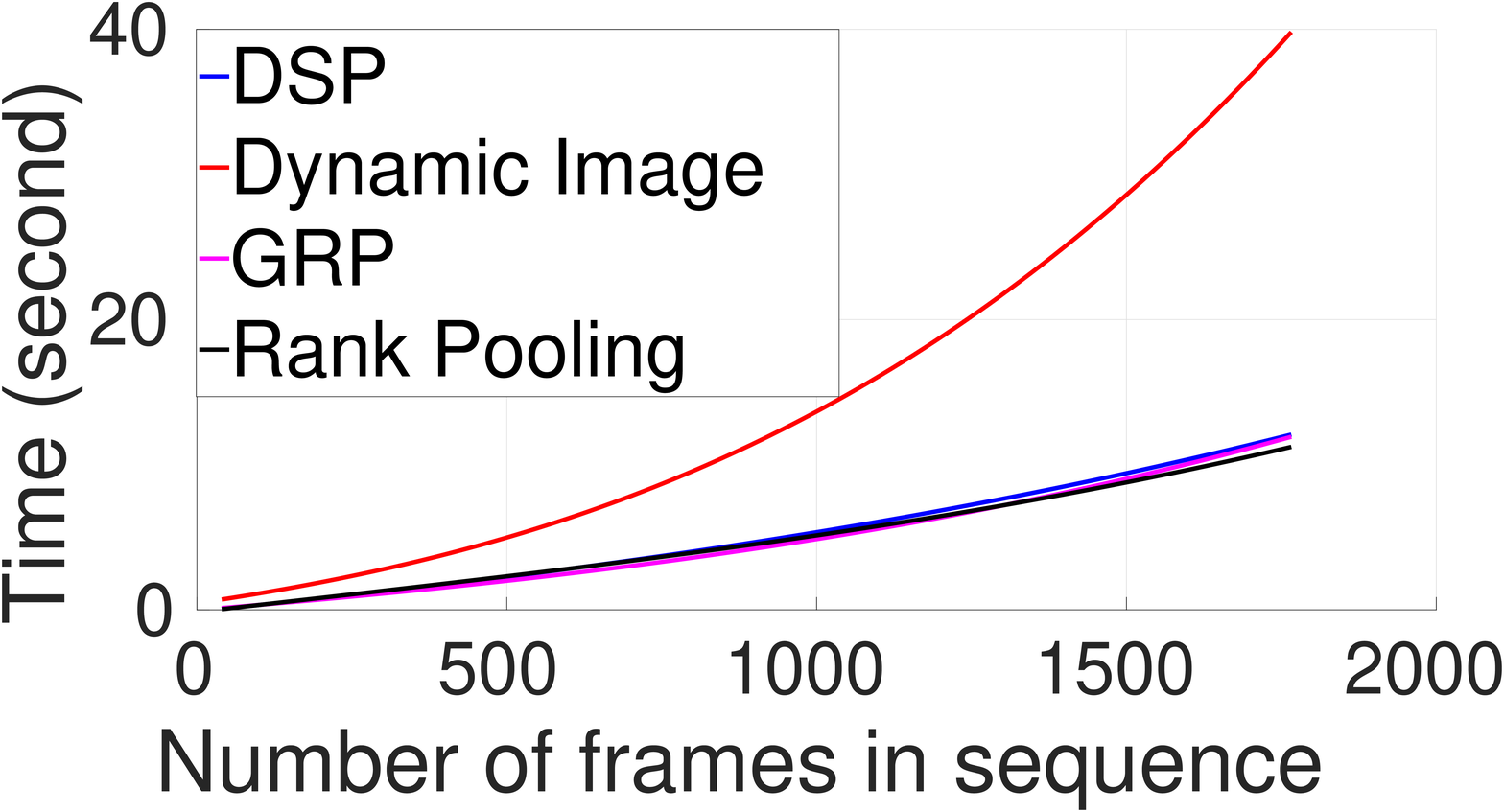} 
}{%
  \caption{Run time analysis of DSP against GRP~\cite{grp}, RP~\cite{fernando2015modeling}, and Dynamic Images~\cite{bilen2016dynamic}.
  \label{fig:runtime}}%
}
\capbtabbox{%
	\begin{adjustbox}{width=1\linewidth}
  \begin{tabular}{c|c|c|c|c|c|c|c}
\hline
\#frames & 1    & 80   & 100  & 140  & 160  & 180  & 260  \\ \hline
\#classes     & 51   & 49   & 34   & 27   & 23   & 21   & 12   \\ \hline
AP~\cite{carreira2017quo}                & \textbf{80.8} & 81.8 & 86.1 & 84.1 & 82.3 & 78.0 & \textbf{77.3} \\ \hline
DSP (ours)              & \textbf{81.6} & 82.8 & 88.5 & 88.0 & 86.1 & 83.3 & \textbf{82.6} 
\label{T2}
\end{tabular}
\end{adjustbox}
}{%
  \caption{Comparison of I3D performance on sequences of increasing lengths in HMDB-51 split-1.}%
}
\end{floatrow}
\end{figure}

\paragraph*{\textbf{Run Time Analysis:}} In Figure~\ref{fig:runtime}, we compare the run time of DSP with similar methods such as rank pooling, dynamic images, and GRP. We used the Matlab implementations of other schemes and used the same hardware platform (2.6GHz Intel CPU single core) for our comparisons. To be fair, we used a single hyperplane in DSP. As the plot shows, our scheme is similar in computations to rank pooling and GRP.


\para{Analysis of Results on I3D Features:} To understand why the improvement of DSP on I3D (80.9\% against our 81.5\%) is not significant (on HMDB-51) in comparison to our results on other datasets, we further explored the reasons. Apparently, the I3D scheme uses chunks of 64 frames as input to generate one feature output. However, to obtain DSP representations, we need a sufficient number of features per video sequence to solve the underlying Riemannian optimization problem adequately, which may be unavailable for shorter video clips. To this end, we re-categorized HMDB-51 into subsets of sequences according to their lengths. In Table~\ref{T2}, we show the performance on these subsets and the number of action classes for sequences in these subsets. As our results show, while the difference between average pool (AP) (as is done in~\cite{carreira2017quo}) and DSP is less significant when the sequences are smaller ($<$80 frames), it becomes significant ($>$5\%) when the videos are longer ($>$260 frames). This clearly shows that DSP on I3D is significantly better than AP on I3D. 

\begin{figure}[!h]	
	\begin{center}
        \includegraphics[width=1\linewidth,trim={0cm 0cm 0cm 0cm},clip]{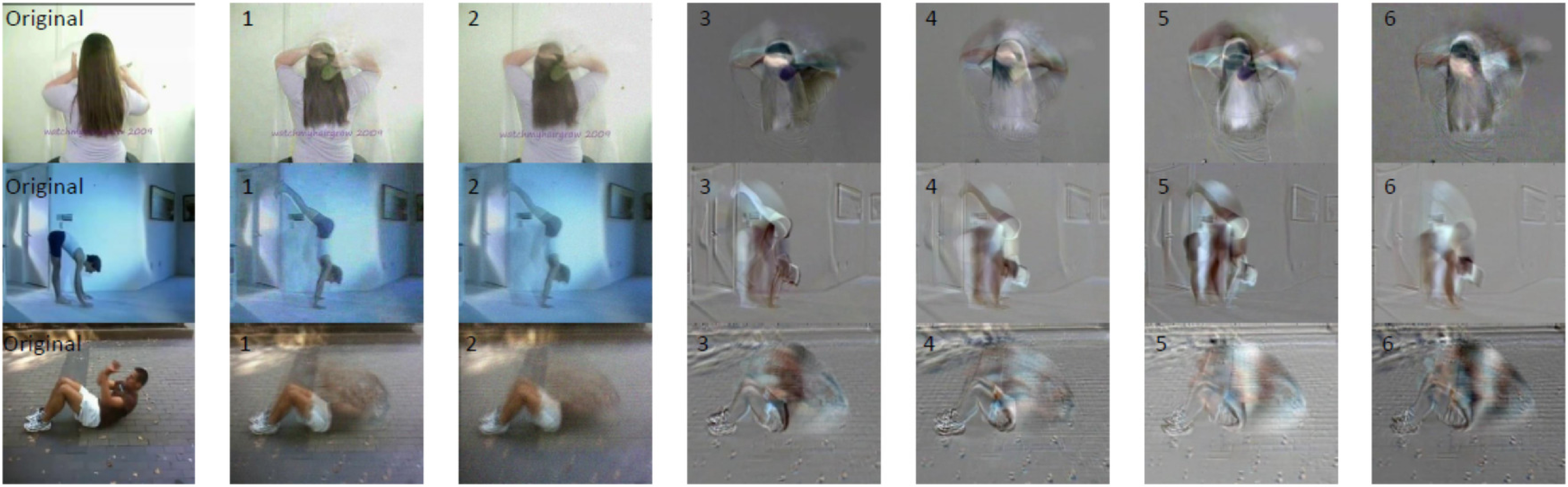}
	\end{center}
	\caption{Visualizations of our DSP descriptor (when applied on raw RGB frames) on an HMDB-51 video sequences. First column shows a sample frame from the video, second-to-seventh columns show the six hyperplanes produced by DSP. Interestingly, we find that each hyperplane captures different aspects of the sequences--first two mostly capture spatial, while the rest capture the temporal dynamics at increasing granularities.}	
    \label{fig:4}
\end{figure}
\para{Qualitative Results:} In Figure~\ref{fig:4}, we visualize the hyperplanes that our scheme produces when applied to raw RGB frames from HMDB-51 videos -- i.e., instead of CNN features, we directly feed the raw RGB frames into our DSP, with adversarial noise generated as suggested in~\cite{moosavi2017universal}. We find that the subspaces capture spatial and temporal properties of the data separately; e.g., the first two hyperplanes seem to capture mostly the spatial cues in the video (such as the objects, background, etc.) while the rest capture mostly the temporal dynamics at greater granularities. Note that we do not provide any specific criteria to achieve this behavior, instead the scheme automatically seem to learn such hyperplanes corresponding to various levels of discriminative information. In the supplementary materials, we provide comparisons of this visualization against those generated by PCA and generalized rank pooling~\cite{grp}. 


\section{Conclusions}
\label{sec:conclude}
In this paper, we investigated the problem of contrastive representation learning for video sequences. Our main innovation is to generate and use synthetic noise, in the form of adversarial perturbations, for building the negative pairs, and then producing our video representation in a novel contrastive pooling scheme. Assuming the video frames are encoded as CNN features, such perturbations are often seen to affect vulnerable parts of the features. Using such generated perturbations to our benefit, we propose a discriminative classifier, in a max-margin setup, via learning a set of hyperplanes as a subspace, that could separate the data from its perturbed counterpart. As such hyperplanes need to fit to useful parts of the features for achieving good performance, it is reasonable to assume they capture data parts that are robust. We provided a non-linear objective for learning our subspace representation and explored efficient optimization schemes for computing it. Experiments on several datasets explored the effectiveness of each component in our scheme, demonstrating state-of-the-art performance on the benchmarks.
\bibliographystyle{splncs}
\bibliography{egbib}
\newpage
\title{Learning Discriminative Video Representations Using Adversarial Perturbations:\\Supplementary Material} 

\titlerunning{Video Representation with Adversarial Perturbations}

\authorrunning{J. Wang and A. Cherian}

\author{Jue Wang\inst{1}\and
Anoop Cherian\inst{2}}

\institute{${^1}$Data61/CSIRO, ANU, Canberra\quad ${^2}$MERL Cambridge, MA \\
	\email{jue.wang@anu.edu.au\quad cherian@merl.com}
}

\maketitle

\section{Discriminative Subspace Pooling: Intuitions}
In the following, we analyze the technicalities behind DSP in a very constrained and simplified setting, that we believe will help it understand better. A rigorous mathematical analysis of this topic is beyond the scope of this paper.

Let us use the notation $X$ to denote a matrix of $n$ data features, let $Z$ be the noise bag, and let $\epsilon$ be the adversarial noise. Then $Z=X+\epsilon$, where $\epsilon$ is fixed for the entire dataset. Simplifying our notation used in the main paper, let us further assume we are looking for a single dimension $w$ that could separate the two bags $X$ and $Z$. Then, this implies for example, in an ideal discriminative setting, $w^TX^i=1, \forall i=1,2,\cdots ,n$ and $w^TZ_i = w^T(X^i+\epsilon) = -1, \forall i=1,\cdots, n$. Substituting the former into the latter, we have $w^T\epsilon = -2$. Combining, we have a set of $n+1$ equations\footnote{Practically, we should use inequalities to signify the half-spaces, but here we omit such technicality.} as follows:
\begin{align}
\label{eq:hype_1}
w^TX_i &= 1,\ i=1,2,\cdots, n\\
w^T\epsilon &= -2.
\label{eq:hype_2}
\end{align}
Assuming $\epsilon$ is a direction in the feature space that could confuse a well-trained classifier, the above set of equations suggest that when learning the discriminative hyperplane in a max-margin setup, our framework penalizes (at twice the rate) directions that could be confusing.

\comment{
Let us further simplify this setting. Suppose we are working in a 2D feature space, where say $x=[x_1, x_2]$ is a data point. Hypothetically, let $x_1$ be 1D a CNN feature capturing the video background, while let $x_2$ captures the dynamics. When learning the noise, suppose that the dimension $x_2$ might be easier to make a classifier to mis-classify (vice versa). Based on this, assuming $\ell_1$ regularization on the noise patterns, the learned universal perturbation $\epsilon$  would be $[0, e]$ (noise will be added on the easily mis-classified dimension) . In this case, our negative bag is given by $z=[x1, x2+e]$. Using the above equations~\eqref{eq:hype_1}~\eqref{eq:hype_2}, assuming $w=[w_1,w_2]$ is the discriminative direction, then we can show that
\begin{equation}
w = [(e+2x_2)/(x_1e), -2/e],
\end{equation}
where the second dimension, which is vulnerable to noise, is made a constant, and the other dimensions undergoes a non-linear transformation with respect to the data dimensions $x_1$ and $x_2$ and the noise $e$ -- the entire setting can be seen as a dimensionality reduction and a non-linear transformation.  
}

\section{End-to-End CNN Learning}
\label{sec:end-to-end}
As alluded to in the main paper, end-to-end CNN training through the discriminative subspace pooling (DSP) layer can be done using methods that are quite well-known. For a reader who might be unfamiliar with such methods, we provide a detailed exposition below. 
\begin{figure*}[t]
\centering
\includegraphics[width=13cm,trim={0cm 8cm 0cm 0cm},clip]{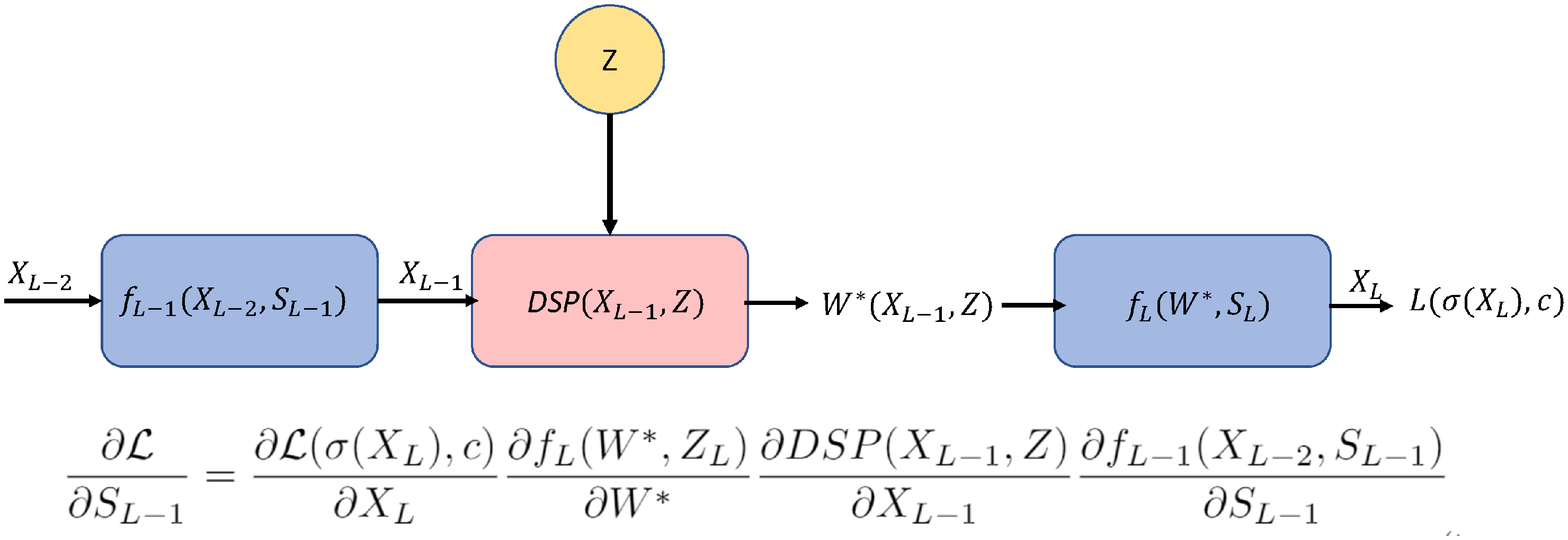}
\caption{Architecture of our end-to-end CNN with discriminative subspace pooling (DSP) layer in between. We assume $X_{\ell}$ represents the feature map outputs from the $\ell$-th CNN layer (from all frames in the sequence) denoted as $f_{\ell}$, and $S_{\ell}$ represents its respective parameters. The final loss is shows as $\mathcal{L}$, $\sigma(\beta)$ is the softmax function, and $c$ is the action class label. The parameter $W$ is the subspace pooled output of the DSP layer, and $Z$ is the adversarial noise. Below the model, we provide the gradient that we are after for enabling back-propagation through the DSP layer.} 
\label{fig:dpl}
\end{figure*}
To set the stage for our discussions, we first provide our CNN architecture with the DSP layer. This CNN model is depicted in Figure~\ref{fig:dpl}. In the model, we assume the DSP layer takes as input the feature map $X_{L-1}$ from the previous layer (across all frames) and the adversarial noise $Z$, and produces as output the subspace descriptor $W^*$. This $W^*$ goes through another series of CNN fully connected layers before using it in a loss layer $\mathcal{L}$ (such as cross-entropy) to be trained against a ground truth video class $c$. Among the gradients of parameters $S$ on the various blocks, the only non-trivial gradient is the one for the block penultimate to the DSP layer, to update the parameters $S_{L-1}$ of this layer will require the gradient of the DSP block with respect to its inputs $X_{L-1}$ (the gradient that we are interested in is depicted below our CNN model in Figure~\ref{fig:dpl}). The main challenge to have this gradient is that it is not with regard to the weights $W$, but the outcome of the DSP optimization $W^*$ -- which is an argmin problem, that is:
\begin{equation}
W^* = \argmin_{W} \dsp(X_{L-1}, Z).
\end{equation}

Given that the Riemannian objective might not be practically amenable to a CNN setup (due to its components such as exponential maps, etc. that might be expensive in a CNN setting), we use a slightly different objective in this setup, given below (which is a variant of Eq. (3) in the main paper). We avoid the use of the ordering constraints in our formulation, to simplify our notations (however we use it in our experiments).
\begin{equation}
\min_{W} \dsp(X) := \Omega(W) + \sum_{i=1}^n \left[\max\left(0, 1-\max\left(y_iW^\top X^i\right)\right)\right]^2,
\label{eq:2}
\end{equation}
where $\Omega(W)=\fnorm{W^TW-\eye{p}}^2$ is the subspace constraint specified as a regularization. Recall that $y_i$ is binary label for frame $i$. With a slight abuse of notation to avoid the proliferation of the CNN layer $L$ in the derivations, we use $X$ to consist of both the data features and the adversarial noise features, as captured by their labels in $y$ ($y=-1$ for adversarial noise features and 1 otherwise), and that the pair $(X^i, y_i)$ denote the $i$-th column of $X$ and its binary label respectively.

\subsection{Gradients for Argmin}

In this section, we derive the gradient $\frac{\partial \dsp(X)}{\partial X}$. We use the following theorem for this derivation, which is well-known as the implicit function theorem~\cite{chiang1984fundamental},~\cite{faugeras1993three}[Chapter 5] and recently reviewed in Gould et al.~\cite{gould2016differentiating}. 
\begin{theorem}
Let $\dsp:\reals{d\times n}\to\reals{d\times p}$ be our discriminative subspace pooling operator on $n$ features each of dimension $d$ (defined as in~\eqref{eq:2}). Then, its gradient wrt $X^i$ is given by:
\begin{equation}
\nabla_{X^i} \dsp(W; X) = -\left.\left\{\nabla_{WW} \dsp(W;X)\right\}^{-1} \nabla_{X^iW} \dsp(W; X^i)\right|_{W=W^*}
\end{equation}
\label{thm:1}
\end{theorem}

The above theorem suggests that to get the required gradient, we only need to find the second derivatives of our objective. To simplify notation, let $P(t,q)$ denote a $d\times p$ matrix, with all zeros, except the $q$-th column which is $t$. Then, for all $i$ satisfying $\max(y_iW^TX^i)<1$, we have the second-order derivatives as follows:
\begin{equation}
\nabla_{WW} \dsp(W; X) = \Omega''(W) +  2\sum_{i} \vecp\left(P\left(\alpha_j(i),j(i)\right)\right)\vecp\left(P\left(\alpha_j(i), j(i)\right)\right)^T,
\end{equation}
where $j(i)=\argmax_q y_i W^TX^i$ and $\alpha_j(i)=y_iX^i$, $q$ capturing the dimension-index that takes the largest of $y_iW^TX^i$, which is a $p\times 1$ vector. Similarly, 
\begin{equation}
\nabla_{X^iW} \dsp(W; X) = 2\vecp\left(P\left(\alpha_j(i),j(i)\right)\right)\vecp\left(P\left(\beta, j(i)\right)\right)^T,
\end{equation}
where $j$ and $\alpha_j$ are as defined above, while $\beta=P(y_iW, j(i))$. Note that $\nabla_{WW}$ is a $pd\times pd$ matrix, while $\nabla_{X^iW}$ is a $pd\times d$ matrix. While, it may seem expensive to compute these large matrices, note that it requires only the vectors $y_i X^i$ as its elements which are cheap to compute, and the argmin takes only linear time in $p$, which is quite small (6 in our experiments). However, computing the matrix inverse of $\nabla_{WW}$ can still be costly. To avoid this, we use a diagonal approximation to it. 

Figures~\ref{subfig:1} and~\ref{subfig:2} show the convergence and the action classification error in the end-to-end learning setup on the HMDB-51 dataset split1 using a ResNet-152 model.
\begin{figure}[h]
	\begin{center}
        \subfigure[]{\label{subfig:1}\includegraphics[width=0.3\linewidth]{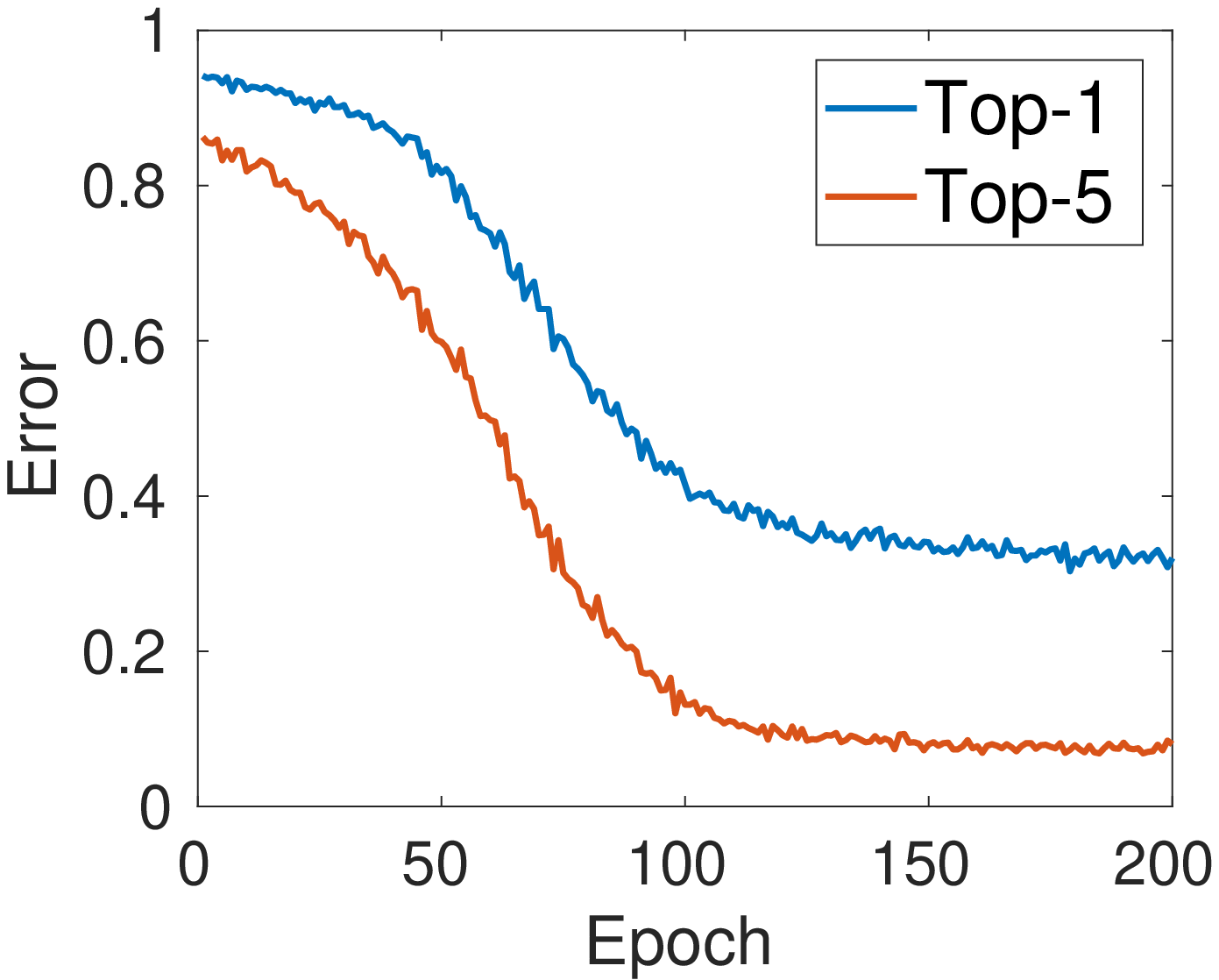}}
        \subfigure[]{\label{subfig:2}\includegraphics[width=0.3\linewidth]{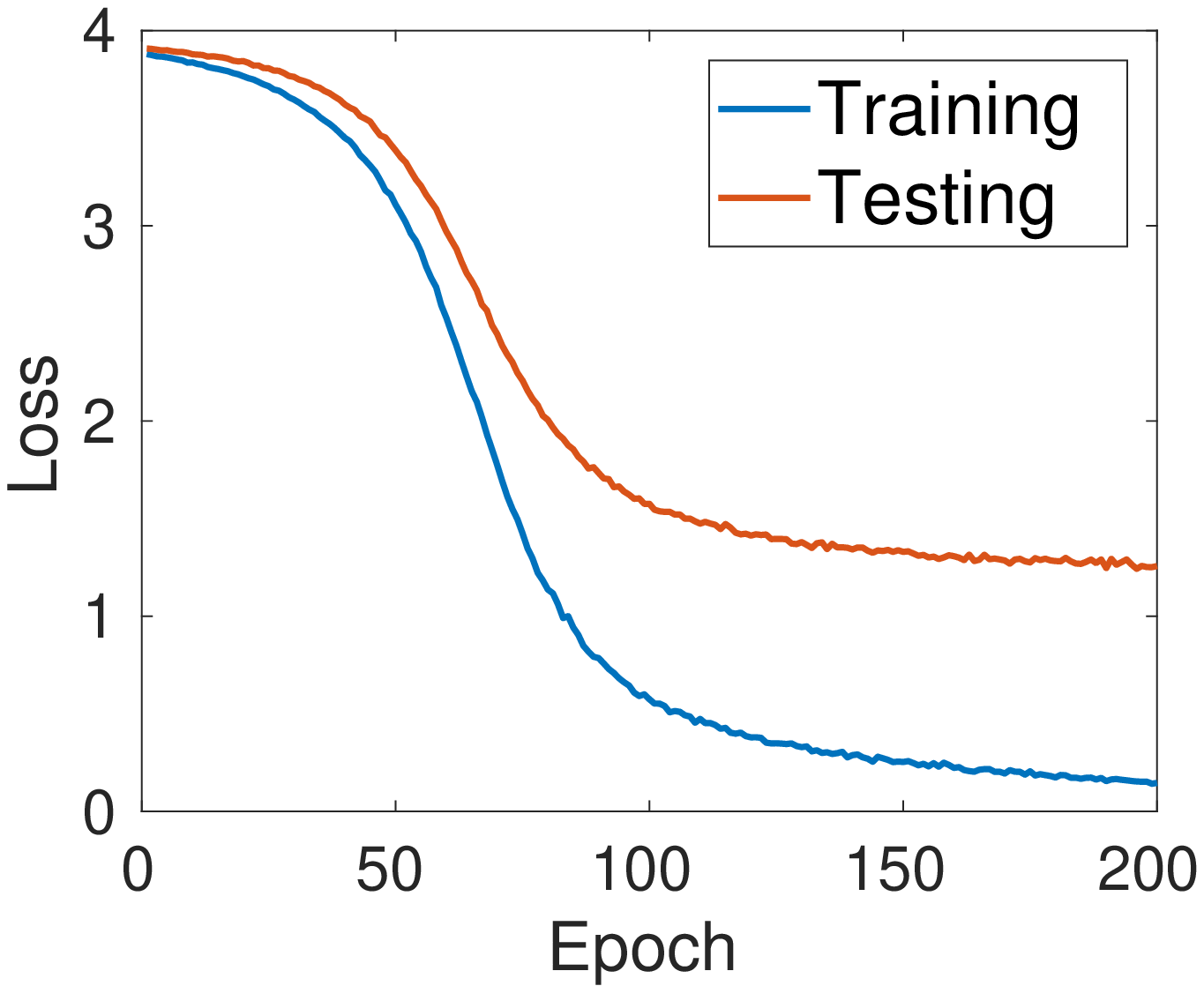}}
	\end{center}
	\caption{Convergence of our end-to-end training setup on HMDB-51 split1.}
\end{figure}

\section{Classifying DSP descriptors Using Neural Networks}
\begin{figure}
\begin{floatrow}
\ffigbox{%
  \includegraphics[width=0.7\linewidth,trim={0cm 0cm 0cm 0cm},clip]{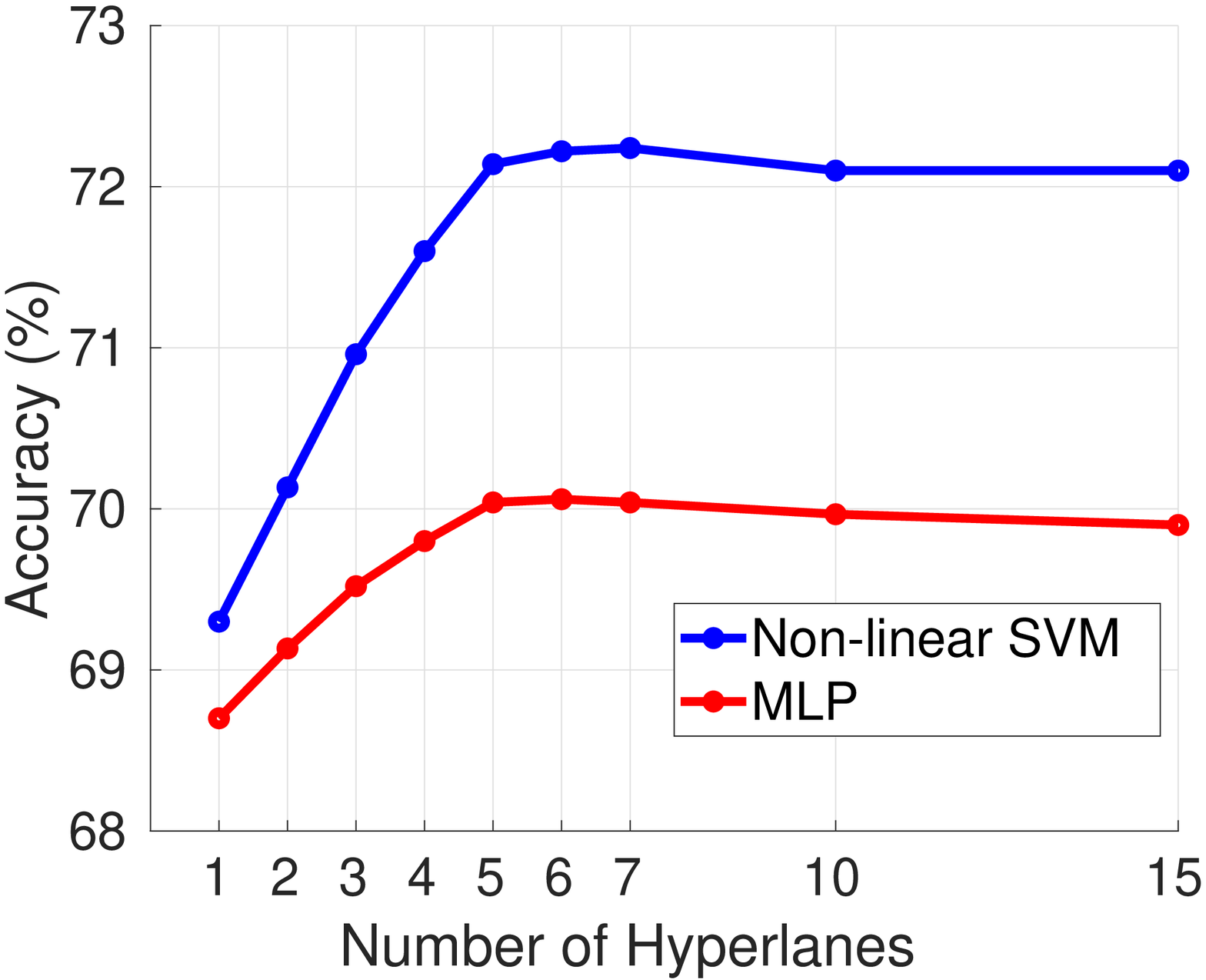}
}{%
  \caption{Accuracy comparisons when using non-linear SVM and multi-layer perceptron for classifying the DSP descriptors (on HMDB-51 split1).
  \label{fig:mlp}}%
}
\capbtabbox{%
	\begin{adjustbox}{width=0.8\linewidth}
  \begin{tabular}{|l|l|l|l|l|l|l|}

\hline
    &RGB & Flow\\ \hline
DSP (SVM)& 58.5     & 67.0     \\ \hline
DSP (end-to-end)& 56.2     & 65.0      \\ \hline
\end{tabular}
\end{adjustbox}
}{%
  \caption{Comparison between end-to-end learning and SVM-based DSP classification on HMDB-51 split-1 with ResNet152.}%
    \label{e2e}
}
\end{floatrow}
\end{figure}

Besides, the above end-to-end setup, below we investigate an alternative setup that mixes frameworks -- that is, use a Riemannian framework to generate our DSP descriptors, and a multi-layer perceptron for video classification. That is, we explore the possibility of training a multi-layer perceptron (MLP) on the discriminatively pooled subspace descriptors, as against a non-linear SVM (using the exponential projection metric kernel) suggested in the main paper. This is because, an SVM-based classifier might not be scalable when working with very large video datasets. In Figure~\ref{fig:mlp}, we compare the performance of this experiment on HMDB-51 split-1. For the MLP, we use the following architecture: we first introduce a $1 \times p$ vector to do a linear pooling of the $p$ columns produced by the DSP scheme. The resultant $d\times 1$ vector is then passed through a $d\times 51$ weight matrix learned against the data labels after a softmax function. We use RELU units after the weight matrix, and use cross-entropy loss.

The result in Figure~\ref{fig:mlp} suggests that the non-linear kernels perform better than using the MLP, especially when the number of hyperplanes $p$ is large. This might be because the $1 \times k$ vector could not capture as much information of each hyperplane as the exponential projection metric kernel does. Note that the subspaces are non-linear manifold objects, and thus the linear pooling operation could be sub-optimal. However, the result of MLP is still better than the baseline result shown in the Table 1 in the main paper. We also provide the result from end-to-end learning setup in the Table~\ref{e2e}, which is slightly lower than the one from SVM setup; which we believe is perhaps because of the diagonal approximations to the Hessians that we use in our back-propagation gradient formulations (see the end of Section~\ref{sec:end-to-end}.

\section{Additional Qualitative Experiment Results}
In the main paper, we make comparison between non-linear kernel on DSP against a linear kernel on AP and MP, which some may argue as unfair, as the latter kernel is expected to me much more richer than the former (and thus it is unclear if the performance in DSP comes from the use of the non-linear kernel or the representation itself). To this end,  we explore the impact of various kernel choices for the methods. Note that the output of the DSP representation is a matrix with orthogonal columns and is an element of the non-linear Stiefel manifold, which to the best of our knowledge, cannot be embedded into a linear space of finite dimensions without distortions. Thus, using a linear SVM may be mathematically incorrect. That said, however, we evaluate the idea in Table~\ref{T1}(left): (i) use a linear classifier (SVM) on DSP and (ii) use non-linear classifier on other features (RBF kernel+SVM). As is clear, linear SVM on DSP is 4-8\% inferior and using non-linear SVM on AP/MP did not improve over DSP -- demonstrating that it is not the classifier that helps, rather it is the DSP representation itself. 
\begin{table}[ht]
\begin{adjustbox}{width=1.0\linewidth}
\centering
\begin{tabular}{|l|l|l|l|l|l|l|}
\hline
    & RGB(L) & Flow(L) & RGB(NL) & Flow(NL)\\ \hline
AP  & 46.7   & 60.0    & 44.2    & 57.8    \\ \hline
MP  & 45.1   & 58.5    & 40.6    & 56.1    \\ \hline
DSP & 50.4   & 63.2    & 58.5    & 67.0    \\ \hline
\end{tabular}
\quad
 \begin{tabular}{|l|l|l|l||l|l||l|l|}
\hline
             & RGB & Flow & R+F & {NTU-S} & {NTU-V} & {YUP-S} & {YUP-M} \\ \hline
RP           & 48.6   & 58.3   & 65.2     & 71.6   & 80.5   & 91.3  & 81.6  \\ \hline
GRP          & 53.3   & 63.4   & 70.9     & 76.0   & 85.1   & 92.9  & 83.6  \\ \hline
DSP          & 58.5   & 67.0   & 72.4     & 81.6   & 88.7   & 95.1  & 88.3  \\ \hline
\end{tabular}
\end{adjustbox}
\caption{Left: Comparison between different classifiers (L)inear and non-linear (NL) on HMDB-51 split-1 with ResNet152. Right:Comparison of DSP against other pooling schemes, esp. rank pooling (RP) and generalized (bi-directional) rank pooling (GRP) on HMDB-51 (two-stream ResNet-152), NTU, and YUP datasets (following the official evaluation protocol)}
\label{T1}
\end{table}

Apart from that, we also make comparison with the Generalized Rank Pooling (GRP) scheme\cite{grp}, which is one of the most important baseline method mentioned in the main paper. Table~\ref{T1}(right) shows the results on all the three datasets. As is seen, DSP still outperforms these prior methods. For RP and GRP, we used the public code from the authors without any modifications. We used 6 subspaces for GRP after cross-validation.

\begin{figure}[!h]
	\begin{center}
    	\subfigure[]{\label{subfig:4}\includegraphics[width=0.32\linewidth]{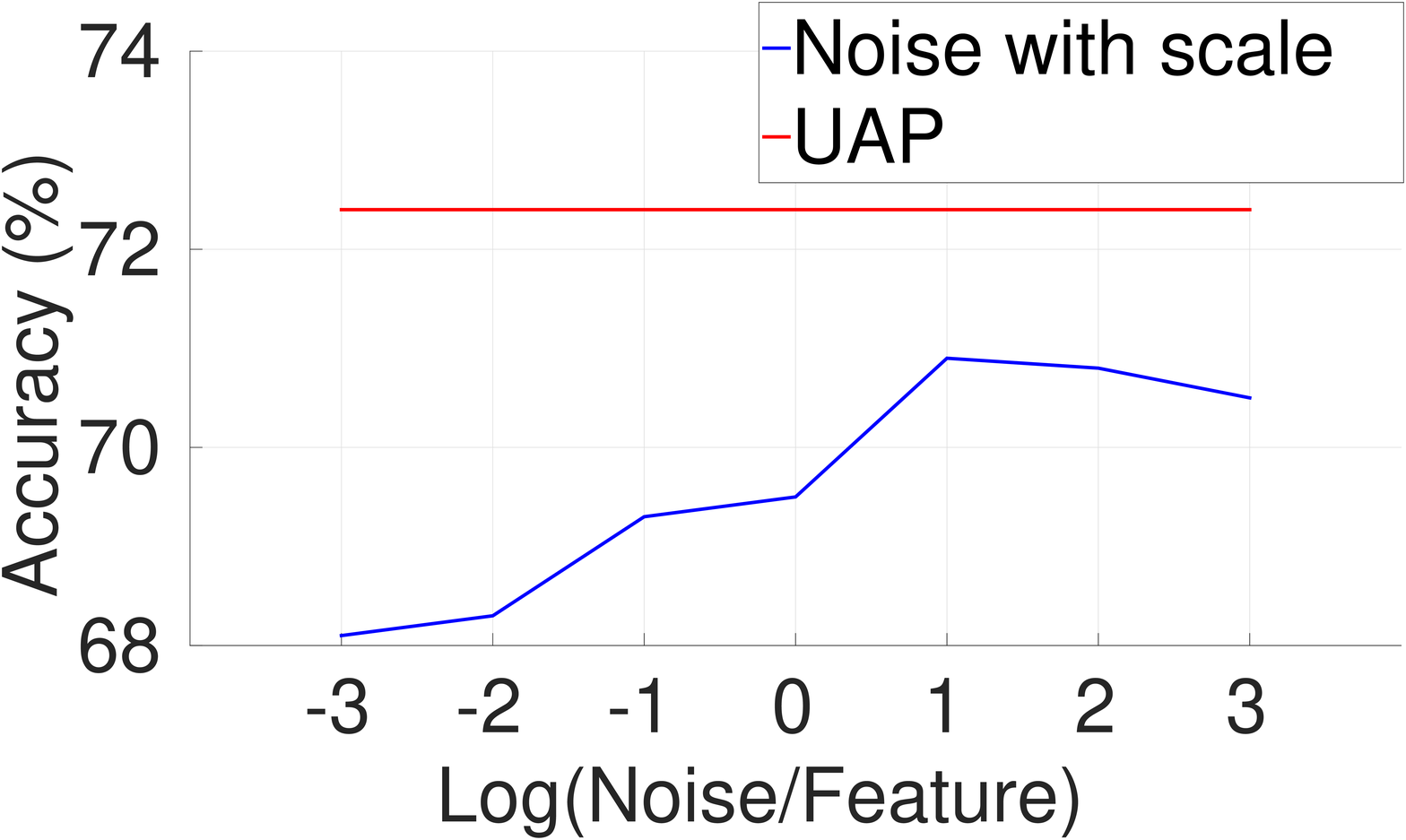}}         
        \subfigure[]{\label{subfig:3}\includegraphics[width=0.32\linewidth]{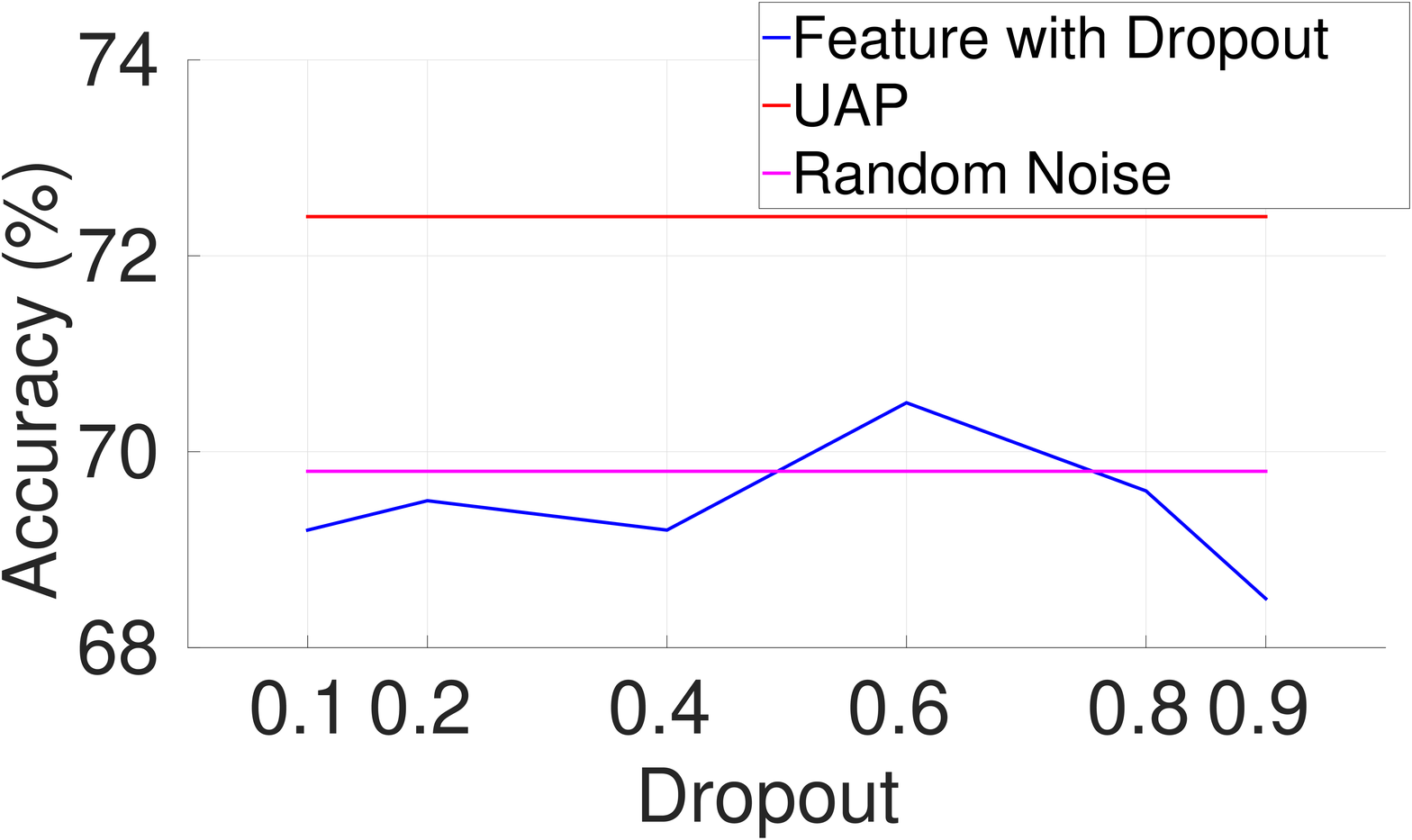}}                      
	\end{center}
	\caption{(a) SNR plot with Gaussian noise and (c) Dropout to generate negative features.}
	\label{fig:all_plots}
\end{figure}
\subsection{More on Noise Selection}
In this section, we explore other alternatives to noise selection, as against the adversarial noise (UAP) we used in the main paper. First, we use different levels of Gaussian noise (signal-to-noise ratio); the results are shown in Figure~\ref{subfig:4}. As the magnitude of noise increases, accuracy do increase, however is below that achieved when using UAP. A second alternative to UAP\footnote{We thank an ECCV reviewer for suggesting this alternative.} is to add drop-out to build the noise bag. The result is provided in Figure~\ref{subfig:3}. Specifically, instead of UAP, we use a negative bag containing features after dropout on the original video features. We increase dropout ratio in the plot, which does improve accuracy, however is below UAP. 


\begin{figure}[]
\begin{center}
	\includegraphics[width=0.9\linewidth,trim={0cm 0cm 0cm 0cm},clip]{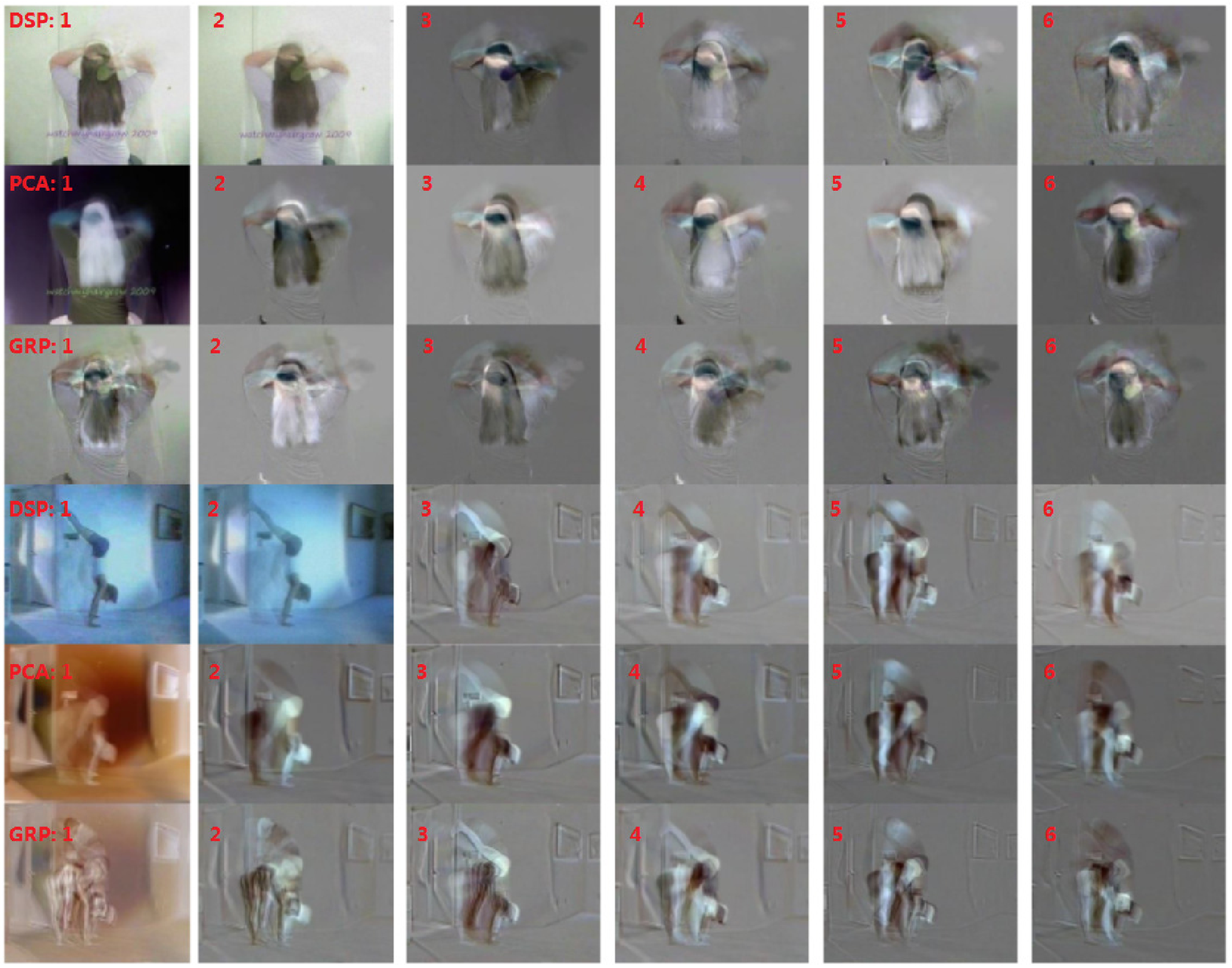}
\end{center}
\caption{We make qualitative comparisons of DSP subspaces against those learned via PCA and GRP~\cite{grp}.}
\label{comparison}
\end{figure}
\subsection{Qualitative Comparisons to PCA and GRP}
In Figure~\ref{comparison}, we make comparison between the visualization of subspaces from DSP, PCA and GRP. From these two set of examples, We find that the type of discriminative subspaces that DSP learns is quite different from the other two. Interestingly, we find that DSP disentangles the appearance and dynamics; however that is not the case in PCA or GRP; even when GRP uses temporal-ordering constraints on top of PCA. 

\end{document}